\documentclass[12pt,authoryear]{article}
\pdfoutput=1

\usepackage{amssymb,amsmath}
\usepackage{mathtools}
\usepackage{acronym}
\usepackage{ccg}
\usepackage{makecell}
\usepackage{booktabs}
\usepackage{placeins}
\usepackage{url}
\usepackage{authblk}
\usepackage{natbib}

\DeclareMathOperator*{\argmax}{argmax}

\begin{document}
\acrodef{mr}[MR]{meaning representation}
\acrodef{ud}[UD]{universal dependency}
\acrodef{ccg}[CCG]{combinatory categorial grammar}
\acrodef{em}[EM]{expectation maximization}
\acrodef{lf}[LF]{logical form}
\acrodef{nlp}[NLP]{natural language processing}
\acrodef{cla}[CLA]{child language acquisition}
\acrodef{cds}[CDS]{child-directed speech}
\acrodef{pos}[POS]{part of speech}

\author[1]{Louis Mahon}
\affil[*]{\textit{lmahon@ed.ac.uk}}
\affil[*]{https://lou1sm.github.io/louismahonville/}
\author[2]{Omri Abend}
\author[2,3]{Uri Berger}
\author[4]{katherine Demuth}
\author[5]{Mark Johnson}
\author[1]{Mark Steedman}

\small{
\affil[1]{School of Informatics, University of Edinburgh}
\affil[2]{Computer Science and Engineering, Hebrew University of Jerusalem}
\affil[3]{School of Computing and Information Systems, University of Melbourne}
\affil[4]{Department of Linguistics, Macquarie University}
\affil[5]{School of Computing, Macquarie University}
}

\title{A Language-agnostic Model of Child Language Acquisition}

\date{}
\maketitle

\begin{abstract}
This work reimplements a recent semantic bootstrapping \ac{cla} model, which was originally designed for English, and trains it to learn a new language: Hebrew. The model learns from pairs of utterances and logical forms as meaning representations, and acquires both syntax and word meanings simultaneously. The results show that the model mostly transfers to Hebrew, but that a number of factors, including the richer morphology in Hebrew, makes the learning slower and less robust. This suggests that a clear direction for future work is to enable the model to leverage the similarities between different word forms.
\end{abstract}

\section{Introduction}
This paper concerns computational models of \ac{cla}, which seek to understand the process of language acquisition by programming a computer to emulate the learning undergone by the child. When presented with data from a given language, such an algorithm should learn a degree of proficiency in that language. The fact that any child, when exposed to appropriate data, can learn any language establishes a strong connection between acquisition and language variation: whatever varies between languages must be specified by the data and must be learnable. It also makes it an essential requirement of a convincing \ac{cla} model that it be capable of learning any language. Here, we reimplement a recent computational \ac{cla} model \citep{abend2017bootstrapping}, which is based on combinatory categorial grammar \citep{steedman2001syntactic} and semantic bootstrapping \citep{pinker1979formal}, and is trained with an expectation-maximization style algorithm. This model is a suitable choice for understanding the acquisition process because of its cognitive plausibility. The dominant paradigm of large language models requires too much training data to be plausible models of how humans acquire language. Even on the small end of the scale they generally train on several orders of magnitude more tokens than a human sees in their entire life. Some have sought to better approximate human learning by learning from a more modest 10-100M tokens \citep{conll-2023-babylm}. However, such models still generally make a number of implausible design choices, such as multi-epoch training, batched parameter updates, and arbitrary text tokenization, and they do not, as we do, ensure that the training examples are presented in the order they appear to the child. Abend et al. (2017) in contrast is grounded in a well-developed theoretical model of semantic bootstrapping, and trains on each example only once, individually, in the order they appear to the child.

We test this model on two languages: English, on which it was originally tested, and Hebrew. The data we use is comprised of real child-directed utterances, taken from the CHILDES corpus \citep{macwhinney1998childes}, coupled with a recent method for converting universal dependency annotations to logical forms \citep{ida2023}.

Firstly, we replicate the findings of \cite{abend2017bootstrapping}, and show that this model is successful in learning the important features of English syntax and semantics. We focus in the present paper on word order learning, and learning the meaning and syntactic categories of individual words. The results show that, after training, the model, correctly, strongly favours SVO order, predicts the right semantics for commonly appearing words and the right syntactic category for most. Then we apply the same training and testing procedure to the Hebrew corpus. There, the model learns word order and word meaning with a reasonably high accuracy. Its accuracy on syntactic categories is somewhat lower than that on English. We then discuss the difference in acquisition performance with respect to the linguistic differences between the two languages, and outline future extensions to the model that can more completely handle the learning of Hebrew without compromising the learning of English. Together these results demonstrate the model in question is broadly successful in transferring between multiple languages, and support the argument that, in general for computational \ac{cla} models, it is important and instructive to evaluate on more than just a single language. The code for training and evaluation will be released on publication.

\section{Method}
\subsection{Theoretical Underpinnings}
Our model deals with syntax and semantics learning only. It assumes the child either has already learned to segment the speech stream and detect potential word boundaries, as evidenced in even young prelinguistic infants \citep{Matt:99}, or is jointly learning phonotactics and morphology with syntax, as in the model of \cite{GoldbergY:13c}. At that point, the child must learn to combine atomic units (words) to produce a meaning representation that depends on (a) the meaning of the constituent words and (b) the manner in which they combine. Initially, for such a child, both are unknown. Theoretically, our approach to this problem falls under Semantic Bootstrapping Theory, 
\citep{pinker1979formal,grimshaw1981form,Brow:73,Bowerman:73b,Schlesinger:71}, which operates as follows: When a child hears an utterance Bambi is home, we assume that, from a combination of perceptual context and background and innate knowledge, it can approximately identify the meaning of the entire utterance as some object in some state: $home(bambi)$. The task is then to figure out which words correspond to which parts of the meaning representation, and the language-specific principles by which they combine. As well as the correct interpretation, where English subjects precede VPs, others are also possible, e.g. Bambi means $home(\cdot)$, is home means $bambi$ and subjects follow VPs. The original implementation of our model \citep{abend2017bootstrapping} designed a language learner that bootstraps learning of both (a) and (b) simultaneously.

\subsection{Combinatory Categorial Grammar.}
\Ac{ccg} \citep{steedman2001syntactic} is a suitable theoretical framework for learning of this sort, because of its tight coupling of syntax and semantics. For each combination of syntactic categories, CCG provides a precise description of a corresponding semantic combination, $bambi + \lambda x. home(x) \rightarrow home(bambi)$. A full example is given  for the two CHILDES \citep{macwhinney1998childes} corpora that we apply our model to: Adam (English), in Figure~\ref{fig:ccg-adam-example}, and Hagar (Hebrew) in Figure~\ref{fig:ccg-hagar-example}.

\begin{figure}[htb]
\centering
    
\deriv{4} {
    \text{you}  & \text{lost}          &      \text{a} & \text{shoe}        \\
    \uline{1}    & \uline{1}            & \uline{1}     & \uline{1}         \\
    \text{NP}    & \text{(S$\bs$NP)/NP}  & \text{NP/N}   & \text{N}          \\
    :you        & :\lambda x.\lambda y. lost\; y\; x         & :\lambda x. a\; x & :shoe                         \\
                 &                      & \fapply{2}                        \\
                 &                      & \mc{2}{\text{NP}}                 \\
                 &                      & \mc{2}{: a\; shoe}          \\
                 & \fapply{3}                                               \\
                 & \mc{3}{\text{S\bs NP}}                                   \\
                     & \mc{3}{:\lambda y. lost\; y\; (a\; shoe)}               \\
                   \bapply{4}                                              \\
                   \mc{4}{\text{S}}                                        \\
                   \mc{4}{: lost\; you\; (a\; shoe)}                                        \\
}
\caption{Example of a CCG derivation for a simple sentence from the Adam (English) corpus.} \label{fig:ccg-adam-example}

\end{figure}

\begin{figure}[htb]
\centering
    
\deriv{3} {
    \text{hu}  & \text{xotek}          &      \text{ec} \\
    \uline{1}    & \uline{1}            & \uline{1}  \\
    \text{NP}    & \text{(S$\bs$NP)/NP}  & \text{NP} \\
    :hu?        & :\lambda x.\lambda y.  xatak\; y\; x    & :ec-BARE \\
                 &                      \fapply{2}                        \\
                 & \mc{2}{\text{S\bs NP}}                                   \\
                     & \mc{2}{:\lambda y. xatak\; y\; ec-BARE}               \\
                   \bapply{3}                                              \\
                   \mc{3}{\text{S}}                                        \\
                   \mc{3}{: xatak\; hu?\; ec-BARE}                                        \\
}
\caption{Example of a CCG derivation for a simple sentence from the Hagar (Hebrew) corpus. The sentence translates to English as ``hes cutting wood'', literally ``he cut-pres-p wood''.} \label{fig:ccg-hagar-example}

\end{figure}
Typically, \ac{ccg} parsing is discussed in terms of combining constituents via combinatory rules to derive a root. For example, the last step of the derivation in Figure~\ref{fig:ccg-adam-example} uses the Backward Application Rule: $Y, X\bs Y \rightarrow X$. Our learning model, when interpreting a sentence-meaning pair, runs these combinators in reverse, that is, it proceeds by successively splitting a root into smaller chunks until they can be aligned with word spans. We will thus often speak of \ac{ccg} `splits', by which we mean the \ac{ccg} combinators run in reverse. The net effect is that our model considers all possible ways to split up the sentence and the meaning representation so that the semantic units best correspond to the language units.

\subsection{Probabilistic Model} \label{subsec:prob-model}
This section describes the details of our new implementation of the semantic bootstrapping CCG-based model of \cite{abend2017bootstrapping}. 

For syntax and semantics learning, the three relevant aspects of each data point are the string of words in the utterance $x$, the meaning representation $m$, and the parse tree $t$. The former two are given by the data, but the parse tree is unobserved, and so treated as a latent variable. We assume that the data is drawn from some joint distribution $P(x,m,t)$, and we fit an approximation to this via several univariate conditional distributions. The conditional distributions are in the generative direction, i.e., together they produce a probability for the word string given the meaning representation. We make the Markovian assumption that the probability of splitting a node depends only on the syntactic category of that node, not on the rest of the tree or on the words or meaning. This means the parse tree can be modelled with a distribution of the form $p_t((s_1,s_2)|s)$, where $s_1$, $s_2$ and $s$ are \ac{ccg} syntactic categories. Note that the tuple $(s_1,s_2)$ is ordered. This distribution also allows the possibility that $s$ is a leaf node, the probability of which is expressed as $p_t(\text{leaf}|s)$. 

We make similar Markovian assumptions on the relationship between syntactic category and meaning, and between meaning and word, so the distribution relating word $x$ and meaning representation $m$ has the form $p_w(x|m)$, and similarly for the distribution relating syntactic category $s$ to meaning representation $m$. Following \cite{abend2017bootstrapping}, we add a second layer of prediction in the form of a shell logical form $e$, between the syntactic category and the logical form. The shell logical form replaces all non-variable terms with a placeholder marked for the function of the placeholder, e.g. verb, entity, determiner. The function is inferred from the CHILDES part-of-speech tag given in the method of \cite{ida2023}. For example the logical form $\lambda y. lost\, y\, (a\, shoe)$ from Figure~\ref{fig:ccg-adam-example} has shell logical form $\lambda y. \text{vconst}\,y\,(\text{detconst}\; \text{nconst})$, because the CHILDES tags for $lost$, $a$ and $shoe$ are `v', `det:art' and `n', respectively. See appendix for full list of conversions from CHILDES tags. This allows the model to share representation power for the structure of the logical form across different examples that may have different values for the constants. It is also used for the measure of word-order preference, described in Section~\ref{subsec:word-order-results}. Thus, $p(m|s)$ is decomposed as $p_l(m|e)$ and $p_h(e|s)$. In $p_h(e|s)$, we ignore slash direction in $s$, so that e.g. conditioning on S\bs NP gives exactly the same distribution as conditioning on S/NP.

Each of these distributions is modelled as a Dirichlet process, to which Bayesian updates are applied at each data point. Taking $p_w$ as an example, the form of the posterior is then
\begin{equation} \label{eq:dp-form}
p_w(x|m) = \frac{n(x,m) + \alpha H(x|m)}{n(m) + \alpha}\,,
\end{equation}
where $n(x,m)$ is the number of times $x$ and $m$ have been observed together in the past, $n(m)$ is the number of times $m$ has been observed in the past, and $H(x)$ is a pre-defined base distribution. An analogous definition holds for $p_l$, $p_h$ and $p_t$. 
The alpha parameter is set to $1$ for all distributions, corresponding to a uniform Dirichlet prior across simplices. During training, we set $\alpha=10$ in $p_t$ to encourage exploration of different syntactic structures, and $\alpha=0.25$ in $p_w$ to produce more confident predicted word meanings, which we find helps stabilize syntax learning.

\subsection{Training Algorithm}
The parameter updates described in Section~\ref{subsec:prob-model} require tracking the occurrences on which two different elements co-occur. For example, in $p_w$, the probability of predicting the logical form $\lambda x. \lambda y. lost\,x\,y$ to be realized as the word `lost' depends on the number of times that logical form and word were observed together during training. Because we do not observe parse trees directly, we instead employ an expectation-maximization algorithm, as follows. When the model observes a single data point $X$, consisting of an utterance as a string and a corresponding logical form, it uses its current parameter values $\theta^{(t)}$ to estimate a distribution over all possible parses that connect the two. The probability assigned to a parse tree $T$ and the data point $X$ is the following product
\begin{gather} \label{eq:evidence}
p(X,T|\theta^{(t)}) = \prod_{s'} p_t(s_1, s_2|s') \prod_{s} 
p_t(\text{leaf}|s)p_h(e_s|s)p_l(m_s|e_s)p_w(x_s|m_s)\,,
\end{gather}
where $s'$ ranges over all non-leaf nodes in $T$, $s_1$ and $s_2$ are the children of $s'$, $s$ ranges over all leaf nodes in $T$, and $e_s$, $m_s$ and $x_s$ are, respectively, the shell logical form, the logical form, and the word aligned to $s$ in $T$.
As we observe $X$, we are interested in the conditional probability of a given parse tree
\begin{equation} \label{eq:bayes-posterior}
p(T|X,\theta^{(t)}) = \frac{p(X,T|\theta^{(t)})}{\sum_{T' \in \mathcal{T}} p(X,T'|\theta^{(t)})}\,,
\end{equation}
where $\mathcal{T}$ is the set of all allowable parses of $X$.
For each parse tree, the co-occurrences that it gives rise to are recorded in proportion to the parse tree's probability. Combining the standard \ac{em} update rule with the Bayesian update for the Dirichlet process, then, for each parameter $\theta$ that tracks the co-occurrence of two elements $a$ and $b$, the update rule is given by
\[
\theta^{(t+1)} = \theta^{(t)} + \mathbb{E}_{T \sim p(T|X,\theta^{(t)})}[\delta_T(a,b)]\,,
\]
where $\delta_T(a,b)$ is an indicator function that is 1 if $a$ and $b$ co-occur in $T$ and 0 otherwise.

The set $\mathcal{T}$ of allowable parse trees is the set of all valid \ac{ccg} parse trees that have the observed \ac{lf} as root, the words in the observed utterance as leaves, and that have congruent syntactic and semantic types.\footnote{We use `semantic/syntactic type' and `semantic/syntactic category' interchangeably.} 

This constraint is based on the assumption that the child knows the semantic type of a \ac{lf} (or fragment thereof on some internal parse node). We approximate this knowledge using the \ac{pos} tags as in the \ac{lf}. The learner uses a mapping (shown in \ref{app:childes-to-semcats}) from these tags to semantic types. The tags were included in the original CHILDES transcription of the utterances, and maintained in the conversion procedure of \cite{ida2023}, which generated our \acp{lf}. For example, the tag `n:prop', indicating a proper noun, gets the semantic category $e$. In the case of ambiguous tags, such as the `v' tag, which does not distinguish between transitive ({<}e,{<}e,t{>}{>}) and intransitive (${<}e,t{>}$) verbs, we associate the node with a set of semantic types, and count it as permissible if its \ac{lf} is congruent with any of these types. This use of CHILDES tags is as a proxy for semantic types of constants in the LF. The learner does not use them to infer part of speech. The full mapping from tags to types is given in \ref{app:childes-to-semcats}. 

Type-raising \citep{steedman2001syntactic} introduces extra lambda variables into the \ac{lf}, therefore, if a node has been type-raised, we count the slashes of the canonical non-type-raised version of its \ac{lf}. 

Adjectives are handled with the special `hasproperty' predicate, which is given semantic category ${<}{<}{<}e,t{>},{<}e,t{>}{>},{<}e,t{>}{>}$, and requires exactly two lambda binders, and nouns are allowed to have no lambda binders and still be counted as permissable. 

Additionally, we require the semantic category to be congruent with the \ac{ccg} category, for each node. \ac{ccg}'s tight coupling of syntax and semantics provides a straightforward mapping from syntactic to semantic categories. In particular, the \ac{ccg} atomic categories $S$ and $NP$ correspond to the Montagovian $t$ and $e$ respectively, and the slashes in non-atomic categories correspond to functions between types. For example, the \ac{ccg} category $S{\bs}NP/NP$ has the semantic type ${<}e, {<}e,t{>}{>}$. See \cite{steedman2001syntactic} for further details. If, when expanding a parse tree, any node violates these constraints, then that branch of search is terminated. 

These constraints are an extension over the original model of \citet{abend2017bootstrapping} They speed up training significantly. We do not have runtime figures for \citeauthor{abend2017bootstrapping}, but based on our experiments, we find the speed-up to be at least 30x. They also make training more robust by removing the noise of updates from inconsistent parse trees. We believe this is the reason for our improved robustness to noise in the \acp{lf}, as described in Section~\ref{subsec:distractors}.

The computation of all allowable parse trees can be performed efficiently by caching the probability of each subtree.

\subsection{Worked Example}
\begin{figure}
    \centering
    \includegraphics[width=\textwidth]{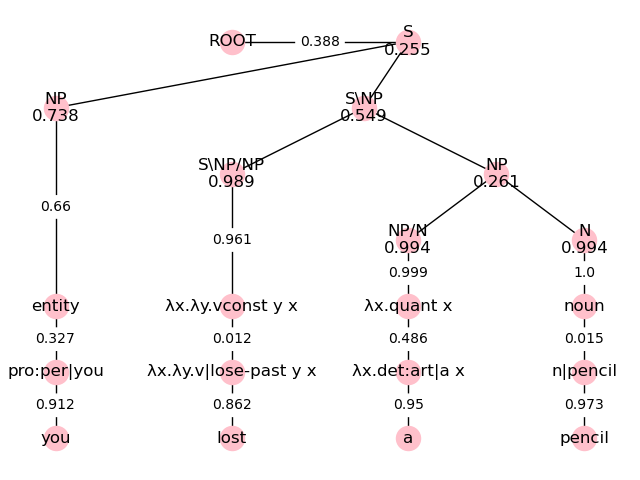}
    \caption{One of the parses considered by the learner for this example. Lambdas are written $L$ and variables are numbers $0, \dots, $. }
    \label{fig:example-parse}
\end{figure}

Here we present a worked example on a single training point. Recall 
that each training example consists of an utterance and corresponding logical form, and the learner considers the set $\mathcal{T}$ of all compatible parses, i.e. all parses with the observed LF as root, the observed utterance as leaves and that obeys the constraints described in Section \ref{subsec:prob-model}. We describe the training updates for a single, correct parse for the example ``you lost a pencil'', as shown in Figure \ref{fig:example-parse}. For the purposes of this example, we show the exhaustive computation for every node in the tree. In practice the probabilities for upper nodes can be cached giving a large increase in efficiency.

The prediction of the tree proceeds from the root. First, the learner predicts a possible root category, here $S$ with probability $0.388$.

Next, it selects a possible split of the root syntactic category $S$, here the selected split is into NP and S\bs NP. 

Then, for the left child node, it predicts the probability for a split into two daughter syntactic categories, or alternatively that this node is a leaf. Here, the prediction is that the node is a leaf, with probability $0.738$. Then it predicts probabilities for a shell logical form given the syntactic category (`entity' with probability $0.66$), a logical form given this shell LF ($you$ with probability $0.327$) and word (or span of words) given this LF (``you'' with probability $0.912$). 

Meanwhile, for the right child of the root, it predicts, with probability $0.549$, a split into further categories of S$\bs$NP/NP and NP. 

Then for the left child of this node, it makes the equivalent predictions it made for the far left node, predicting a split or leaf given the syntactic category (`leaf' with probability $0.989$), then of shell LF given syntactic category ($\lambda x. \lambda y. vconst\; x\; y$ with probability $0.961$), of LF given shell LF ($\lambda x. \lambda y. lose_{past}\; x\; y$ with probability $0.012$) and word span given LF (``lost'' with probability $0.862$). 

For the right child of this node, the right-most NP in Figure \ref{fig:example-parse}, it predicts a split into NP/N and N, with probability $0.261$. 

For the NP/N daughter, it predicts `leaf' with probability $0.994$, then a shell LF given syntactic category ($\lambda x. quant\; x$ with probability $0.999$), of LF given shell LF ($\lambda x. \lambda y. det:art|a; x$ with probability $0.486$) and word span given LF (``a'' with probability $0.95$). 

Finally for the N daughter node it predicts `leaf' with probability $0.994$, $noun$ with probability $1.0$ (note these figures are rounded to three places), $n|pencil$ with probability $0.015$, and ``pencil'' with probability $0.973$. 

At this point, the semantic types of each node can be inferred. The rightmost $n|pencil$ node, in virtue of its $n$ CHILDES pos tag, is inferred to have semantic type ${<}e,t{>}$. The $\lambda x.\; det{:}art|a\; x$ node to its left has type ${<}{<}e,t{>},e{>}$. The parent of these two nodes then gets type ${<}e{>}$. Meanwhile the $\lambda x.\lambda y.v|lose_{past}\; x\; y$ has two allowable types based on the $v$, tag, ${<}e,t{>}$ and ${<}e,{<}e,t{>}{>}$. Only the latter is compatible with it having two lambda binders, so the former is discarded. Together with its sibling ${<}e{>}$ node, this gives its parent type ${<}e,t{>}$. Finally, together with its sibling, which is the leftmost leaf that has the tag `pro:per' and so gets the semantic type ${<}e{>}$, these give the parent, which is the root node, type ${<}t{>}$. For all of these nodes, the semantic type is compatible with number of lambda binders and the \ac{ccg} category, therefore the tree is used for training.

The total probability for this tree and leaves given the root LF is then computed by multiplying all the above probabilities:
\begin{gather*} 
0.388 \times 0.738 \times .66 \times .327 \times .912 \times .549 \times .961 \times .012 \times .862 \times \\
\times .261 \times .994 \times .999 \times .486 \times .95 \times .994, \times 1.0 \times .015 \times .973 = 5.281e-7\,.
\end{gather*}

Given that we observe the leaves, we condition on this event by diving by the sum of the probabilities of all elements of $\mathcal{T}$. Here, that sum turns out to be $5.888e-7$, so the conditional probability for the tree in Figure \ref{fig:example-parse} is
\[
\frac{5.281e-7}{5.888e-7} \approx 0.896\,.
\]

Thus, in the Dirichlet processes that are used to make all model predictions, we update the counts of the co-occurrences in this tree by $0.896$. An example of a co-occurrence in this tree is of the LF $you_{pro:per}$ with the word span ``you''. This is shown in the bottom left of Figure \ref{fig:example-parse}. The number of `times' $you_{pro:per}$ has been observed to co-occurr with ``you'' is increased by $0.896$.

This same procedure is repeated for every element of $\mathcal{T}$. In practice, for the corpora we use, this is generally $50-100$ trees. 

\section{Results}
\subsection{Datasets}


In addition to the straightforward SVO examples in Figures~\ref{fig:ccg-adam-example} and~\ref{fig:ccg-hagar-example}, the \acp{lf} can express more complex syntactic features such as modals, negation, prepositional phrases, and relative clauses. Table \ref{tab:lf-examples} shows some more complex examples. Further detailed examples can be found in \cite{ida2023}, as well as full details of the process by which they were produced, and the rationale behind the various design choices involved in this production.

\begin{table}[]
    \centering
    \small
    \begin{tabular}{c|c}
        \textbf{utterance} & \textbf{LF} \\
        \hline
         they 're in the drawer upstairs & ($upstairs\; (in\; (the\; drawer)\; they)$) \\
         penguins can't fly & ($\neg can\; (fly\; penguin_{pl})$  \\
         what are you giving them for dinner ? & $\lambda wh. for\; dinner\; (giving\; you\; wh\; them)$ \\
         get a kleenex and wipe your mouth & $get\; \_\; (a\; kleenex) \& wipe\; \_\; (your\; mouth)$
    \end{tabular}
    \caption{Examples of more complex \acp{lf} that appear in our datasets.}
    \label{tab:lf-examples}
\end{table}

We post-process this data to exclude words that serve only as discourse markers and do not appear in the \acp{lf} or receive the CHILDES part-of-speech tag `co', meaning `communicator', which includes most instances of words like `so', `well' and the child's name. This results in 19314 tokens and 5320 utterances for Adam (English) and 6187 tokens and 3295 utterances for Hagar (Hebrew). As each data point contains exactly one utterance (as well as the corresponding LF), these are also the numbers of data points in each dataset. 

\subsection{Word Order} \label{subsec:word-order-results}
Following the procedure of \cite{abend2017bootstrapping}, we measure the acquisition of grammar and lexicon by examining the model's implicit word order parameters as a proxy. The degree to which the model favours each of the six possible word orders is determined by (a) the probability it assigns to the two CCG splits that are necessary to parse a simple transitive sentence under that order, and (b) the probability it assigns to the respective order in which the subject and object combine with the verb under that order. We assume that the verb-medial orders, SVO and OVS, must combine with the object first, so the six different possibilities are measured as follows:

\begin{align*}
    &p(\text{SOV}) \vcentcolon= \\
    & p_{t}(\footnotesize{\text{(NP,\ S\bs NP)}}|\footnotesize{\text{S}})p_{t}(\footnotesize{\text{(NP,\ S\bs NP\bs NP)}}|\footnotesize{\text{S\bs NP}})p_{h}(\lambda x.\lambda y. \text{vconst $y$ $x$}|\footnotesize{\text{(S\bs NP\bs NP)}}) \\[10pt]
    &p(\text{SVO}) \vcentcolon= \\
    & p_{t}(\footnotesize{\text{(NP,\ S\bs NP)}}|\footnotesize{\text{S}})p_{t}(\footnotesize{\text{(S\bs NP/NP,\ NP)}}|\footnotesize{\text{(S\bs NP)}})p_{h}(\lambda x.\lambda y. \text{vconst $y$ $x$}|\footnotesize{\text{(S\bs NP/NP)}}) \\[10pt]
    &p(\text{VSO}) \vcentcolon= \\
    & p_{t}(\footnotesize{\text{(S/NP,\ NP)}}|\footnotesize{\text{S}})p_{t}(\footnotesize{\text{(S/NP/NP,\ NP)}}|\footnotesize{\text{S/NP}})p_{h}(\lambda x.\lambda y. \text{vconst $x$ $y$}|\footnotesize{\text{(S/NP/NP)}})\\[10pt] 
    &p(\text{OSV}) \vcentcolon= \\
    & p_{t}(\footnotesize{\text{(NP,\ S\bs NP)}}|\footnotesize{\text{S}})p_{t}(\footnotesize{\text{(NP,\ S\bs NP\bs NP)}}|\footnotesize{\text{S\bs NP}})p_{h}(\lambda x.\lambda y.\text{vconst $x$ $y$}|\footnotesize{\text{(S\bs NP\bs NP)}})\\[10pt] 
    &p(\text{OVS}) \vcentcolon= \\
    & p_{t}(\footnotesize{\text{(S/NP,\ NP)}}|\footnotesize{\text{S}})p_{t}(\footnotesize{\text{(NP,\ S/NP\bs NP)}}|\footnotesize{\text{(S/NP)}})p_{h}(\lambda x.\lambda y.\text{vconst $y$ $x$}|\footnotesize{\text{S/NP \bs NP})}\\[10pt] 
    &p(\text{VOS}) \vcentcolon= \\
    & p_{t}(\footnotesize{\text{(S/NP,\ NP)}}|\footnotesize{\text{S}})p_{t}(\footnotesize{\text{(S/NP/NP,\ NP)}}|\footnotesize{\text{S/NP}})p_{h}(\lambda x.\lambda y. \text{vconst y x}|\footnotesize{\text{(S/NP/NP)}})\,. \\ 
\end{align*}
So, for example, the probability of SOV is product of the probability of splitting an $S$ into $NP$ and $S\bs NP$, the probability of further splitting the resulting $S \bs NP$ into $NP$ and $S \bs NP \bs NP$, and the probability of this $S \bs NP \bs NP$ node having a logical form that takes two arguments and places the first in the object position and the second in subject position. (We use the convention that the argument immediately to the right of the verb is the subject.) The third term is what distinguishes $p(\text{SOV})$ from $p(\text{OSV})$.

\begin{figure}
    \centering
    \includegraphics[width=\textwidth]{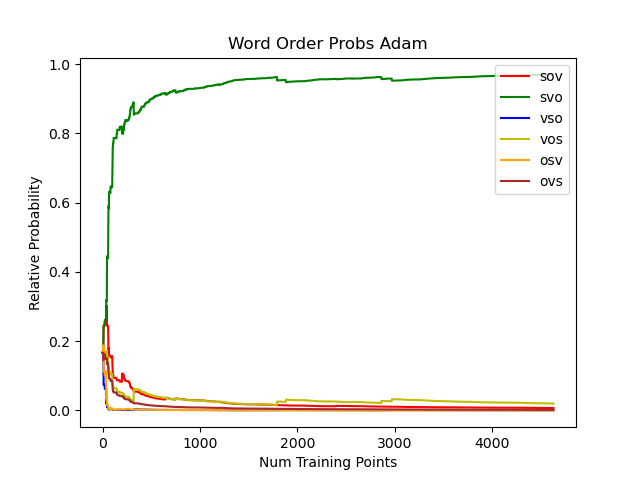}
    \caption{Evolution, over the course of training, of the learner's preference for each of the six possible word orders on Adam (English). It learns rapidly and confidently to favour SVO.}
    \label{fig:adam-word-order-learning}
\end{figure}

\begin{figure}
    \centering
    \includegraphics[width=\textwidth]{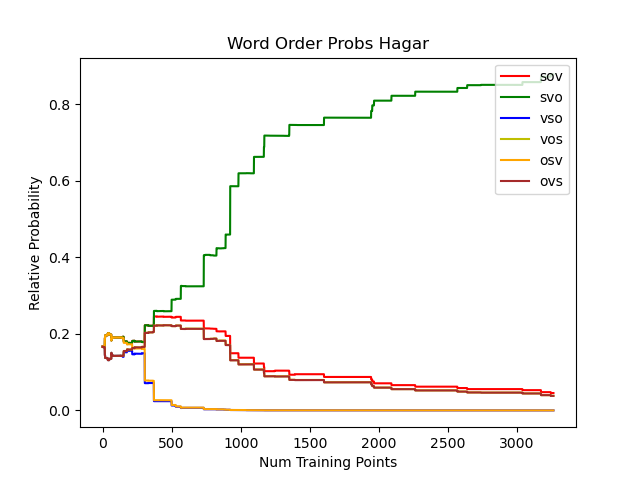}
    \caption{Evolution, over the course of training, of the learner's preference for each of the six possible word orders on Hagar (Hebrew). It learns SVO confidently, but more gradually than on Adam (English), and there are visible jumps where the learner encountered data points that were key for syntax learning.}
    \label{fig:hagar-word-order-learning}
\end{figure}

Figures~\ref{fig:adam-word-order-learning} and~\ref{fig:hagar-word-order-learning} take the relative values of these six scores, by normalizing so they sum to 1, and show how the above measure of word order preference changes over the course of training, for Adam (English) and Hagar (Hebrew) respectively. In both cases, the model succeeds in learning the correct SVO order, but this is faster and more extreme in Adam (English). In Hagar (Hebrew), the learning curve also appears somewhat step-shaped, with the SVO probability jumping up at a number of points, rather than increasing smoothly. 

\begin{figure}
    \centering
    \includegraphics[width=\textwidth]{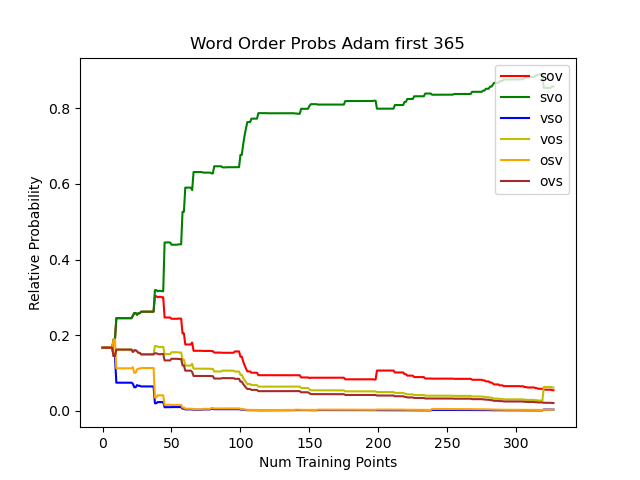}
    \caption{Zoomed in version of the first 365 data points in Adam (English), which contains the same number of critical examples as the full Hagar (Hebrew) dataset. The overall shape is similar to that of Hagar (Hebrew).}
    \label{fig:adam365-word-order-probs}
\end{figure}

We hypothesize that, when learning an ordered category for an unknown word such as a transitive verb, the majority of early learning takes place in a small number of critical examples in which the phenomenon in question is clearly attested, and the child has already learned all the other words. In this case, we should therefore expect a noticeable rise in the probability of the correct order (SVO) on data points which (a) include a transitive meaning representation, (b) have no complicating features such as prepositional phrases, adverbials, reduplication or repetition, and (c) contain only words that the child has already encountered on previous data points. Tracking the number of such points, we find that there are 391 for Adam and 14 for Hagar. In Figure~\ref{fig:adam365-word-order-probs}, we plot learning over just the first 365 points of Adam, as that also gives exactly 14 critical examples. Comparing Figure~\ref{fig:hagar-word-order-learning} (full Hagar dataset) and Figure~\ref{fig:adam365-word-order-probs} (first 365 data points of Adam), which both contain the same number of critical examples, we see that they both produce roughly the same shaped learning curves.

\begin{table}
    \centering
    \begin{tabular}{c|c|c|c|c}
        dataset & \thead{num. critical examples} & \thead{total num. \\ word repeats} & \thead{percentage \\ new words} & \thead{Zipf \\ coefficient} \\
        \hline
        Adam (English) & 391 & 13744 & 7.99 & 1.436\\
        Hagar (Hebrew) & 14 & 3880 & 21.94 & 1.566\\ 
    \end{tabular}
    \caption{Comparison of the diversity of words forms between the two datasets.}
    \label{tab:word-form-diversity}
\end{table}

\begin{figure}
    \centering
    \includegraphics[width=0.8\textwidth]{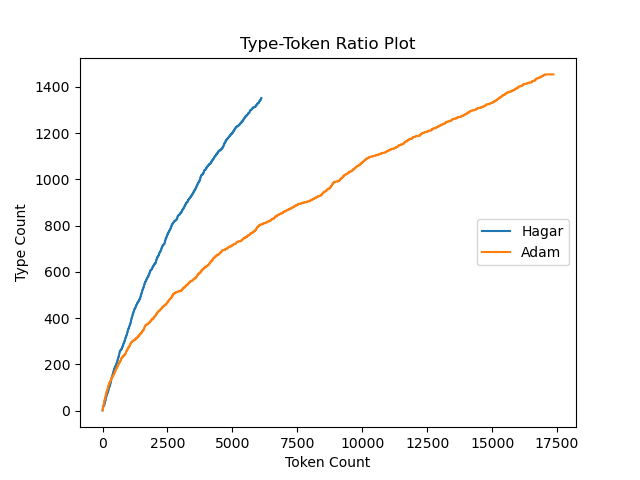}
    \caption{The number of unique types encountered throughout training. Hagar encounters more types for the same number of tokens, consistent with a greater diversity of word forms.}
    \label{fig:type-token-plot}
\end{figure}

The apparent difference in the number of critical examples between Adam and Hagar is explained in part by the fact that the richer morphology leads to more diverse word forms and hence fewer examples when all words have been previously seen. This is measured explicitly, in various ways, in Table~\ref{tab:word-form-diversity}. As well as the count of critical examples, it shows (column 2) the total number of times any word repeats (column 2), the percentage of all words in all utterances that have not been seen before (column 3), and the Zipf coefficient (column 4).\footnote{The Zipfian distribution, which intuitively expresses that a small number of words account for the majority of word occurrences, with a long sparse tail of rare words, has been observed to well-model many corpora, including \ac{cds} \citep{lavi2023zipfian}. Formally, the occurrence frequency of the $n$th most common word is of the form $f_n = \frac{1}{(n+b)^a}$, where $a$ and $b$ are corpus-specific parameters, with $a$ indicating the degree of sparsity and $b$ determining the $y$-intercept. The Zipf coefficient we report is the $a$-parameter after fitting a function of this form to predict $f_n$ from $n$.} For all four measures, we can see that Hagar (Hebrew) has more unseen words. Figure~\ref{fig:type-token-plot} evinces the same property graphically, by plotting the number of unique tokens (types) encountered as a function of the total number of tokens encountered. That is, the plot passes through point $(x,y)$ if and only if, after having seen exactly $x$ tokens, the model has seen exactly $y$ unique tokens. The steeper rise for Hagar (Hebrew) shows that it encounters more unseen word forms.

As well as the greater diversity of word forms, the less smooth learning curve for Hagar (Hebrew) in Figure~\ref{fig:hagar-word-order-learning} is also an artefact of the fact that we have chosen, following \citet{abend2017bootstrapping}, to report the measurement of SVO described in Section \ref{subsec:word-order-results} as a proxy for degree of learning of the grammar as a whole. Although Hebrew is classified as an SVO language, like English, this proxy gives a misleading impression of the course of learning

Firstly, in English, adjective predication (thats dangerous) and statements of membership (hes a man) and identity (thats Daddy) use a copula, providing evidence for SVO structure. The LFs for these sentences are `hasproperty(that,dangerous), equals(that, (a man)) and equals(that, Daddy), respectively, so the model can easily learn that two meaning representations for is are lambda x.lambda y.hasproperty y x lambda x.lambda y.equals y x. Hebrew, on the other hand, expresses these meanings without copulae. For example: ``?at mecunenet''--literally you sick, or ``ze cnonit'', meaning ``this is a small radish''--literally this small-radish. Copular sentences are highly frequent in both corpora.

Relatedly, many examples in the Hagar corpus are one- or two-word utterances, 73.9\% vs 15.2\% for Adam and, of course, sentences of fewer than 3 words cannot exhibit full SVO structure. The most common utterance in Hagar is nakon, meaning ``right/correct'', which alone accounts for 10.2\% of utterances. 
Finally, Adam is a larger dataset: 5320 vs 3295 data points. 

We stress that these differences are not evidence for Hebrew being more difficult to learn in general than English. They mean only that the specific feature of SVO order that we are using as a proxy for overall learning is more strongly attested in English than Hebrew, as the two languages are represented in our datasets of Adam and Hagar. This is consistent with the idea Hebrew is, compared with languages such as English, less strongly committed to SVO structure \cite{Doron:00b}.

\subsection{Word Meaning and Syntactic Category} \label{subsec:lexicon-results}
Going beyond this emergent favouring of SVO word order, what we are ultimately interested in learning is the lexicon, which relates words to pairs of syntactic categories and meaning representations. To evaluate this, we measure the model's prediction for logical form and syntactic category. For each dataset, we select the 50 most common words, and annotate them with a ground-truth logical form and syntactic category. The full CCG lexicon could contain many possibilities for each word, but we restrict to those that are attested in our corpora. See appendix for full list.

We then extract a predicted logical form $m'$ for each word $x$ as follows:
\begin{align}
    m' =& \argmax_m P(m|x) = \argmax_m \frac{P(x|m)P(m)}{P(x)} = \argmax_m P(x|m)P(m) \approx  \notag \\
    &\approx \argmax_m p_w(x|m)p_w(m) = \argmax_m p_w(x,m)\,, \label{eq:lf-pred}
\end{align}
where the last quantity is determined by a Dirichlet process and so can be approximated by the observed number of times that $w$ and $m$ co-occur (recall this is in fact the sum of the expected values of their co-occurrence across all data points).

Similarly, we extract a predicted syntactic category for each word as follows:
\begin{align}
    s' =& \argmax_s P(s|x) = \argmax_s \frac{P(x|s)P(s)}{P(x)} = \argmax_s P(x|s)P(s) =  \notag \\ 
    =& \argmax_s \sum_m \sum_e P(x,m,e|s)P(s) \approx \sum_m \sum_e p_w(x|m)p_l(m|h)p_h(e|s)p_{syn}(s)\,, \label{eq:syn-pred}
\end{align}
and again, the last quantity can be computed straightforwardly, this time as a product of terms from each of the model`s Dirichlet processes. 

\begin{table}[]
    \centering
    \begin{tabular}{c|c|c|c}
        Corpus &  meaning correct & syntactic category correct & both correct \\
        \hline
        Adam (English) &  100\% & 76\% & 76\% \\
        Hagar (Hebrew) &  100\% & 46\% & 34\% \\
    \end{tabular}
    \caption{Accuracy of the learned word meanings and syntactic categories on the fifty most common words, with respect to the manually annotated ground truth.}
    \label{tab:word-acc}
\end{table}

Table~\ref{tab:word-acc} reports the percentage of points for which the predicted meaning representation, as per \eqref{eq:lf-pred} and the predicted syntactic category, as per \eqref{eq:syn-pred} agree with the manually annotated ground truth. For both datasets, the model achieves 100\% on the meaning representation, meaning it pairs every word with the correct logical form. The syntactic category accuracy is lower, 76\% for Adam (English) and 46\% for Hagar (Hebrew)\footnote{Note that the accuracy for both can be lower than that for syntactic category, even with 100\% meaning accuracy. This is because it is possible for the model to get both syntactic category and meaning right, but not get the pair right, for example, if it predicts `NP: $\lambda x. run\; x$', then it is mixing up the syntactic category for run (noun) with the meaning representation for run (verb).}. This reflects the fact that the process of extracting the syntactic category is more involved than that for meaning. The reason the model can predict the correct meaning but the wrong syntactic category is that it first predicts a distribution over the former, and then uses this distribution to predict the latter, as in \eqref{eq:syn-pred}. If the model's prediction goes wrong at the second stage, then the meaning will be correctly predicted but the syntactic category will not.\footnote{In future work we plan to evaluate the course of learning more fully as the acquisition as a grammar and parsing model, incrementally training on weeks $1, \dots, n$ and testing on week $n+1$, using precision and recall of meaning representations as a measure, following \cite{kwiatkowski2012probabilistic}.}

\subsection{Distractor Settings} \label{subsec:distractors}
To test the robustness of word-order learning to noise, we follow Abend et al. and consider a setting in which an utterance is paired not with a single logical form representing its meaning, but with multiple logical forms, only one of which represents its true meaning. The learner is then free to consider any of these LFs as the meaning of the utterance. Formally, what this means is that parse trees are computed for each of these LFs, and all of them are placed in the set $\mathcal{T}$. Then, when calculating the Bayesian posterior, as per \eqref{eq:bayes-posterior}, the denominator is larger, so the probability on any given tree is smaller, as compared to the no-distractor setting. 

This makes learning more difficult, and simulates the fact that there may be some uncertainty for the child as to the meaning a given utterance represents. When there is a single tree that the model is very confident in, then the probability from this tree dominates anyway, and overall there is little effect from the distractor trees. However, when there is no such single confident interpretation, the distractor trees significantly reduce the probability on the trees from the correct LF, including the correct tree, and so dilute the learning effect.  

The other, distractor, logical forms are taken from the utterances immediately following and preceding the given utterance. Specifically, the $n$ distractor setting takes the $\lfloor n/2 \rfloor$ previous examples and the $\lceil n/2 \rceil$ following examples.

For example, in Adam, data points 226-228 are as follows:

Data point 226:
you blow it--$blow\; you\; it$

Data point 227:
you can blow--$can\; (blow\; you)$

Data point 228:
you do it--$do\; you\; it$

Thus, in the two distractor setting, when training on data point 227, we include the parse trees from all three of these LFs. In this case, one possible interpretation takes the LF from data point 226blow(you,it)--and interprets you as meaning $pro{:}per|you$, can as meaning $pro{:}per|it$, blow as meaning $\lambda x.\lambda y.v|blow\; y\; x$, and the sentence as being in SOV order. However, by this stage in training, the model has already learnt to place very little probability on the splits required for SOV, in particular splitting S\bs NP as NP + S\bs NP\bs NP, so this incorrect interpretation has a small probability and doesnt affect training much.

\begin{figure}
    \centering
    \includegraphics[width=0.65\textwidth]{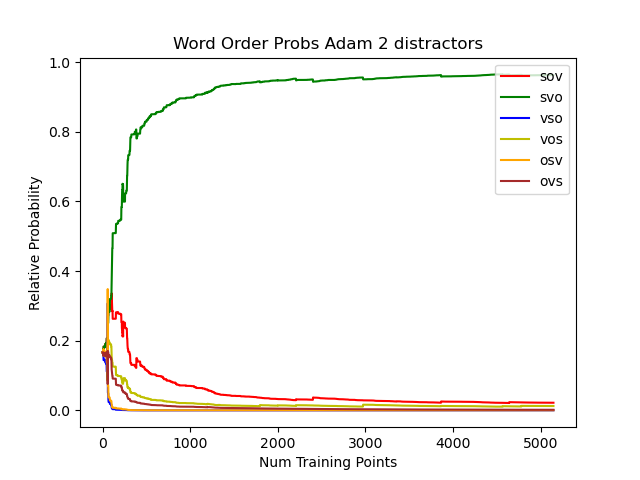}
    \includegraphics[width=0.65\textwidth]{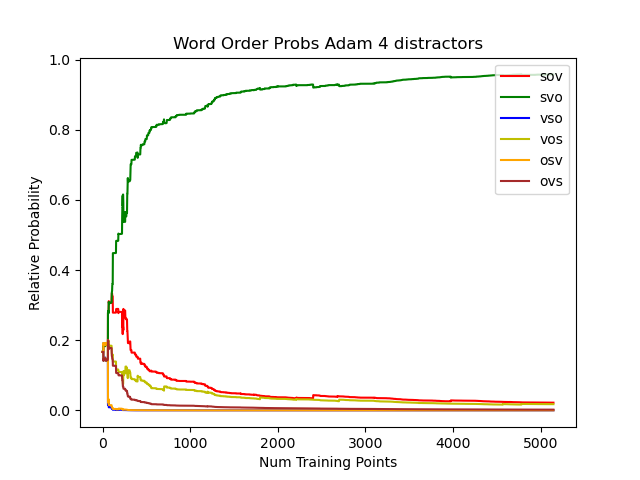}
    \includegraphics[width=0.65\textwidth]{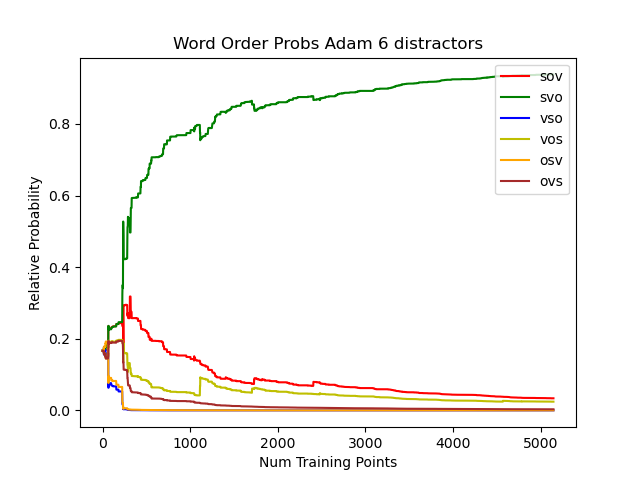}
    \caption{Word-order learning in Adam (English) with different numbers of distractor logical forms. More distractors slightly slows down learning, but the model still succeeds in confidently learning SVO.}
    \label{fig:adam-distractor-word-order-learning}
\end{figure}

As shown in Figure~\ref{fig:adam-distractor-word-order-learning}, for Adam (English), the learner is still capable of learning the correct SVO order in all distractor settings. This is an improvement upon the version in \cite{abend2017bootstrapping}, where dealing with distractor settings required the introduction of an extra `learning rate' parameter, that had to be set to different values for different numbers of distractors. In Appendix~\ref{app:higher-distractors-adam}, we present results for higher numbers of distractors, and show that, on Adam, the learner can handle up to 12 before performance starts to substantially degrade. 

\begin{figure}
    \centering
    \includegraphics[width=\textwidth]{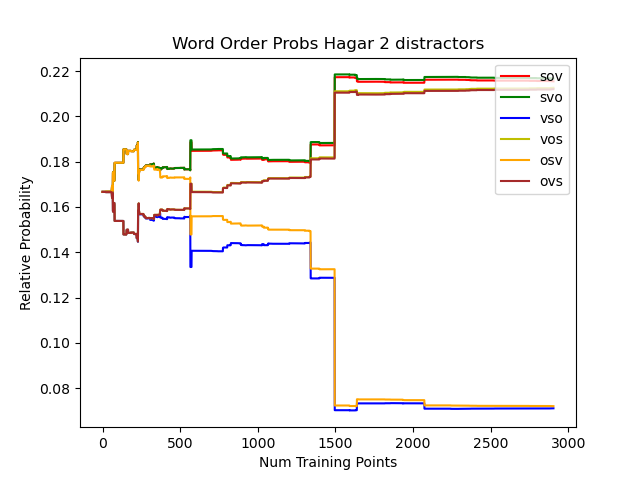}
    \caption{Word-order learning in Hagar (Hebrew) with two distractor logical forms. This prevents the learner from settling on SVO order, which does not rise to high probability and is only very marginally favoured over two of the other five orders.}
    \label{fig:hagar-distractor-word-order-learning}
\end{figure}

For Hagar (Hebrew), however, as shown in Figure~\ref{fig:hagar-distractor-word-order-learning}, the model is not yet able to handle the distractor settings and fails to learn SVO. This fits with the picture, outlined above, that word-order is learnt correctly on Hagar (Hebrew), though currently not with as much confidence or robustness as on Adam (English).

Table~\ref{tab:distractor-word-acc} shows the performance in the distractor settings in terms of word meaning and syntactic category accuracy, as presented in Section~\ref{subsec:lexicon-results}. For both corpora, the accuracy drops as the number of distractors increases. Notably, the difference between the two corpora is much less striking than for word order learning: Adam (English) with six distractors is comparable to Hagar (Hebrew) with two distractors for word meaning and syntactic category learning, but, as seen by comparing Figures~\ref{fig:adam-distractor-word-order-learning} and~\ref{fig:hagar-distractor-word-order-learning}, it is much more successful at the presented measure of word order learning. This again suggests that the difference between Figures~\ref{fig:adam-distractor-word-order-learning} and~\ref{fig:hagar-distractor-word-order-learning} is largely a result of SVO order (as it is measured here and in \cite{abend2017bootstrapping}) being especially strongly attested in English, rather than the learner failing to acquire Hebrew in general.  

\begin{table}[]
    \centering
    \begin{tabular}{c|c|c|c}
        Corpus &  meaning correct & syntactic category correct & both correct \\
        \hline
        \thead{Adam (English) \\ 0 distractors} &  100\% & 76\% & 76\% \\
        \thead{Adam (English) \\ 2 distractors} &  92\% & 62\% & 56\% \\
        \thead{Adam (English) \\ 4 distractors} &  88\% & 42\% & 38\% \\
        \thead{Adam (English) \\ 6 distractors} &  78\% & 32\% & 28\% \\
        \hline
        \thead{Hagar (Hebrew) \\ 0 distractors} &  100\% & 46\% & 34\% \\
        \thead{Hagar (Hebrew) \\ 2 distractors} &  86\% & 36\% & 36\% \\
    \end{tabular}
    \caption{Extension of Table~\ref{tab:word-acc}: word meaning and syntactic category accuracy in distractor settings.}
    \label{tab:distractor-word-acc}
\end{table}

\section{Discussion}

Our approach differs theoretically from other recent approaches to language acquisition. \cite{ambridge2020against} argues that language acquisition can be understood purely based on the recall of all past occasions on which an utterance was used. This is claimed to adequately account for a range of linguistic phenomena without recourse to syntactic or semantic abstractions. Our model, on the other hand, uses abstractions in the form of lexical entries for words, and combinatory \ac{ccg} rules. \cite{chater2018language} treat language acquisition as the learning of a perceptuo-motor skill. One point of emphasis for \cite{chater2018language} is the fact that much relevant information to language learning is forgotten quickly, necessitating that learning occurs rapidly and in real-time (in this sense, the polar opposite of \cite{ambridge2020against}). Another point of emphasis is the social context in which the child hears the utterance. We account for the first point by training on each example only once, one at a time, in the order they appear to the child. Pragmatic context is not currently represented in our input to the learner.

We also differ from these works in that ours is not purely theoretical but is based on a working model. An earlier work, \citet{regier2005emergence} proposes a programmable model whose framework is similar to the theoretical account of \cite{ambridge2020against}. The data consists of utterances paired with manually created binary strings, where each bit indicates the presence or absence of a syntactic feature. \cite{yang2002knowledge} proposes a probabilistic language acquisition model that assumes that the child begins with access to all possible grammars, which can be specified by a finite set of parameters, i.e. the principles and parameters framework \citep{Chom:81,Hyam:86}. It then learns to place more weight on those grammars that successfully parse observed sentences. This differs from our model, which acquires a statisical model of language-specific syntax, lexicon and logical form simultaneously by semantic bootstrapping from utterance meaning representations.

In the realm of word learning from speech, \cite{rasanen2019computational} presents a model for early word learning from real multimodal data, testing on English only.

Some neural bilingual \ac{cla} models have been proposed, which mostly focus on word-meaning learning, and do not attempt the more difficult task of learning from complete sentences. \cite{li2022computational} provide a summary of recent neural models for the related task of learning two languages simultaneously. The present study is, to our knowledge, the first cross-language evaluation of a computational semantic-bootstrapping model of how a child acquires language syntax and semantics.

As evidenced by our results, taking a model originally designed in the context of one language, and testing it on another, can reveal shortcomings which were obscured by the peculiarities of the first language. The main instance of this is the failure of the model to recognize similarities between word forms, which showed up in the learning of syntactic categories for individual words. There is a significantly lower accuracy in predicting Hebrew syntactic categories than English, though the predictions are still often correct. Further analysis reveals the lower accuracy to be largely due to Hebrew`s richer morphology producing a greater diversity of word forms. The model does not recognize these different forms as bearing any similarity, and instead, must learn each independently. This means it encounters fewer utterances where it already knows all word meanings and on the basis of which it can learn syntactic structure.

Testing on a second language also shows which aspects of the model are robust. In our case, the learning of word meanings transfers with high accuracy. Given that it was originally designed when only English data was available, and originally tested on only English data, this is a significant strength of the model. The learning of word order lies somewhere between the very successful learning of word meaning, and the less accurate learning of word syntactic category. The model does still confidently learn the correct SVO dominant order, but it is less robust to noise in the logical forms, and has a more jagged learning curve, and lower final confidence in SVO. This lower performance is noteworthy, but not insurmountable. Although Hebrew morphology is more complex than English, it has been shown to be learnable, by e.g. \cite{GoldbergY:13c}.

Hebrew is a suitable language to use as a first comparison to English, because the two have many, but not all, features in common. 

\section{Limitations}
One limitation of our learner is that it does not model anything below the token level, the tokens being taken from the data of \citet{ida2023}, which in turn took them from the CHILDES parses. The issue highlighted above of the sparsity of word forms suggests a future extension to allow it to guess a meaning for new words if they are similar in form to familiar words. This requires it to learn some internal structure to these tokens in virtue of which they can be similar or dissimilar to one another. One way to do this would be by explicitly adding morphology, and allowing the parse tree to extend not to word boundaries but to morpheme boundaries. Another option would be to add a neural word predictor. A character-level language model could be used to produce vectors for each word that depend on their structure, which could then be fed into a multi-layer perception. Different inflected forms of the same root should then have a similar word vector and so a similar predicted meaning. Note, this approach does not mean a replacement of anything that is currently in the model, it would model only morphology, not syntax or semantics. It may require more data but potentially be more robust to the variety of inflected forms. Either of these additions could be learned independently of the model described here or in conjunction.

This need for the learner to discern a similarity between different inflected forms is obscured by the very sparse morphology in English. Different thematic roles for the same word or nominal phrase are not distinguished by morphological case as they are in Hebrew, so the
model does not need to treat them as separate lexical entries. This further highlights the value of testing computational \ac{cla} models on multiple languages, and future work includes testing on further languages in addition to English and Hebrew.  

Another limitation concerns our method of evaluating our model. Measuring the relative preference for different word orders allows comparison with \citet{abend2017bootstrapping}, but could give a misleading result in certain contexts where the correct analysis is a non-standard word order, e.g. in topicalization.
In future, we hope to adopt a richer and more diverse evaluation suite, including measuring the fraction of test utterances with the correct inferred root \ac{lf} and parse tree.

Thirdly, a more thorough evaluation of our learner involves testing on a more diverse set of languages, with larger and more comparable corpora. In contrast to the two corpora we use here, which differ in the number of tokens and utterances. It would also be interesting to compare different corpora for different children within the same language.

\section{Conclusion}
This paper reimplemented a recent computational model for child language acquisition, based on semantic bootstrapping, which learns from real transcribed child-directed utterances paired with annotated logical forms as meaning representations. We replicated the original results from this model on English, and performed the same evaluation on Hebrew. The results show that the ability of the model generally transfers well to a new language, but its learning on Hebrew is slower and less robust than on English. Further analysis reveals this, not surprisingly, to be due in large part to the richer morphology in Hebrew producing a more diverse set of word forms.
Future work includes the extension of the model to detect and leverage similarities between word forms, application to other languages, and testing on corpora with equal numbers of utterances and tokens.  

\section{Acknowledgements}
This research was supported by ERC Advanced Fellowship GA 742137
SEMANTAX and the University of Edinburgh Huawei Laboratory.

\bibliographystyle{elsarticle-harv} 
\bibliography{bibliography}

\newpage
\appendix

\section{Conversion from CHILDES POS Tags to Shell LF Terms}
As described in Section~\ref{subsec:prob-model}, we use the CHILDES part of speech tags, which are included in the logical forms of \cite{ida2023}, to choose the marking on the constant in the shell logical form. Table~\ref{tab:conversion-chiles-shell} gives full correspondence. In the main text in Section~\ref{subsec:word-order-results}, we indicated the marking with the first letter of the right column, e.g. `verb' gives `vconst'.

\begin{table} 
\caption{Our mapping from CHILDES part of speech tags of terms in the logical form to the marking on the constant in the corresponding shell logical form.} \label{tab:conversion-chiles-shell}
\centering
\resizebox{!}{0.6\textwidth}{
\begin{tabular}{ll}
\toprule
CHILDES TAG & const marking in shell LF \\
\midrule
adj & adj \\
adv & adv \\
adv:int & adv \\
adv:tem & adv \\
aux & aux \\
conj & connect \\
coord & connect \\
cop & cop \\
det & quant \\
det:art & quant \\
det:dem & quant \\
det:int & quant \\
det:num & quant \\
det:poss & quant \\
mod & raise \\
mod:aux & quant \\
n & noun \\
n:pt & noun \\
n:gerund & entity \\
n:let & entity \\
n:prop & entity \\
neg & neg \\
prep & prep \\
pro:dem & entity \\
pro:indef & entity \\
pro:int & WH \\
pro:obj & entity \\
pro:per & entity \\
pro:poss & quant \\
pro:refl & entity \\
pro:sub & entity \\
qn & quant \\
v & verb \\
\bottomrule
\end{tabular}
}
\end{table}

\section{Mapping from CHILDES POS Tags to Montagovian Semantic Types} \label{app:childes-to-semcats}
Table~\ref{tab:mapping-childes-to-semcat} shows how we infer the Montagovian semantic type from the CHILDES POS tags that are available in our \acp{lf}. Some are defined schematically, the avoid overly long expressions. For example, the category for conjunctions (conj) and coordinations (coord) are use the variable $X$ to stand for any other semantic category. The reason the mapping from tags to semantic types is many-to-one is that this allows learning to be shared across categories. For example, if the model learns that the general category det precedes nouns, it knows that this is true for all types of determiners, whereas if we distinguish between det:art, det:poss, det:num etc., then it has to learn this separately for each.

\begin{table} 
\caption{Our mapping from CHILDES part of speech tags of terms in the logical form to Montagovian semantic types.} \label{tab:mapping-childes-to-semcat}
\centering
\resizebox{!}{0.6\textwidth}{
\begin{tabular}{ll}
\toprule
CHILDES TAG & const marking in shell LF \\
\midrule
adj & {<}{<}e,t{>},{<}e,t{>}{>}\\
adv & not considered \\
adv:int & not considered \\
adv:tem & not considered \\
aux & not considered \\
conj & {<}X,{<}X,X{>}{>} \\
coord & {<}X,|{<}X,X{>}{>}\\
cop & handled separately \\
det & {<}{<}e,t{>},e{>} \\
det:art & {<}{<}e,t{>},e{>} \\
det:dem & {<}{<}e,t{>},e{>} \\
det:int & {<}{<}e,t{>},e{>} \\
det:num & {<}{<}e,t{>},e{>} \\
det:poss & {<}{<}e,t{>},e{>} \\
mod & {<}{<}{<}e,t{>},{<}e,t{>}{>},{<}e,t{>}{>} \\
mod:aux & {<}{<}e,t{>},e{>} \\
n & {<}e,t{>} \\
n:pt & {<}e,t{>} \\
n:gerund & e \\
n:let & e \\
n:prop & e \\
neg & {<}{<}e,{<}e,t{>}{>},{<}e,{<}e,t{>}{>}{>},\;  {<}{<}e,t{>},{<}e,t{>}{>}  {t,t} \\
prep & prep \\
pro:dem & e \\
pro:indef & e \\
pro:int & e \\
pro:obj & e \\
pro:per & e \\
pro:poss & {<}e,t{>} \\
pro:refl & e \\
pro:sub & e \\
qn & {<}e,t{>} \\
v & {<}e,{<}e,t{>}{>},\; {<}e,t{>} \\
\bottomrule
\end{tabular}
}
\end{table}

\section{Zipf Plots for Adam (English) and Hagar (Hebrew)}
\begin{figure}
    \centering
    \includegraphics[width=0.75\textwidth]{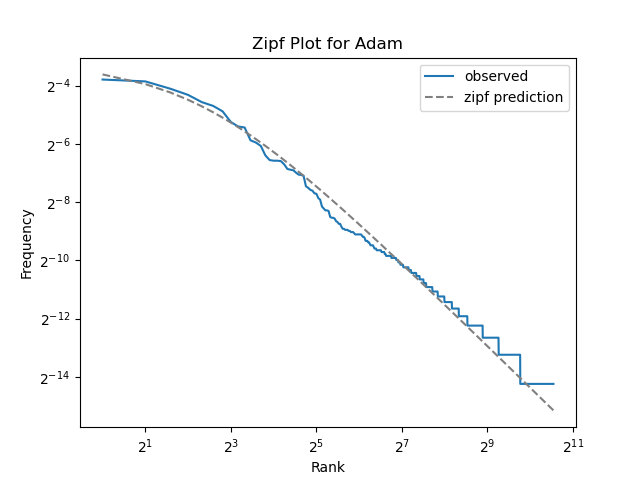}
    \includegraphics[width=0.75\textwidth]{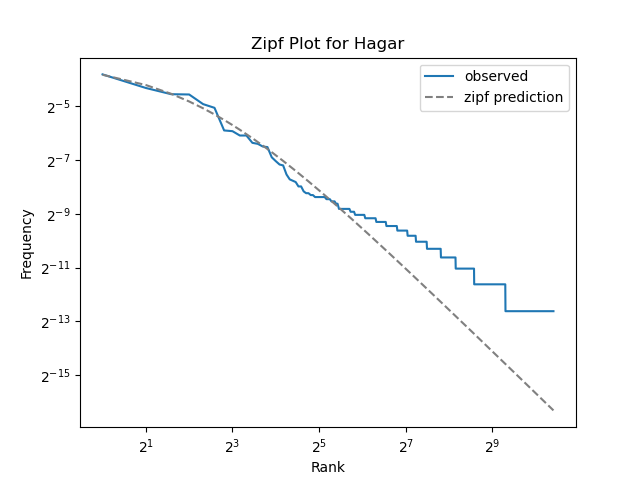}
    \caption{Zipf plots for Adam (English) and Hagar (Hebrew), using the fit $a$ and $b$ parameters. Section~\ref{subsec:word-order-results} reports $a$ as a measure of sparsity.}
    \label{fig:zipf-plots}
\end{figure}

Section~\ref{subsec:word-order-results} reported the Zipf coefficient for the Adam (English) and Hagar (Hebrew) corpora. Here, Figure~\ref{fig:zipf-plots} plots the word frequency against rank, both as observed in the data and as predicted by the fit Zipf function.

\FloatBarrier

\section{Logical Form and Semantic Category Accuracy by Word}
Here we our ground truth annotation, and the learners' predictions, for each of the fifty most common words. The accuracy scores reported in Table \ref{tab:word-acc} refer to the fraction of these words for which the prediction agrees with the the ground trtuh annotation.

\subsection{Manually Annotated Lexicon for Fifty Most Common Words} \label{app:annotated-lexicon}
This section shows the ground-truth logical form meaning representation and \ac{ccg} syntactic category for the fifty most common words in each dataset. As described in Section~\ref{subsec:lexicon-results}, these are used to evaluate the learner's ability to acquire the correct lexicon. Note, the \acp{lf} that appeared in the main paper were abbreviated for clarity. Here, we write the full \ac{lf}, including the CHILDES part of speech tag. The full lexical entry is of the form \mbox{<\ac{lf}> || <syntactic-category>}. Where a word has two common meanings, we include two different lexical entries, separated with a comma.
\vspace{2ex}

\footnotesize
\textbf{Adam}
\begin{verbatim}
'll:\lambda x.\lambda y.mod|~will (x y) || S\\NP/(S\\NP)
're:\lambda x.\lambda y.v|hasproperty y x || S\\NP/NP,\lambda x.\lambda y.v|equals y x || S\\NP/NP
's:\lambda x.\lambda y.v|equals y x || S\\NP/NP,\lambda x.\lambda y.v|hasproperty y x || S\\NP/NP
Adam:n:prop|adam || NP
I:pro:sub|i || NP
a:\lambda x.det:art|a x || NP/N
an:\lambda x.det:art|a x || NP/N
another:\lambda x.qn|another x || NP/N
are:\lambda x.\lambda y.v|equals x y || S\\NP/NP,\lambda x.\lambda y.v|hasproperty y x || S\\NP/NP
break:\lambda x.\lambda y.v|break y x || S\\NP/NP
can:\lambda x.\lambda y.mod|can (x y) || S\\NP/(S\\NP),\lambda x.\lambda y.mod|can (x y) || S/NP/(S\\NP)
d:\lambda x.\lambda y.mod|do (x y) || S\\NP/(S\\NP),\lambda x.\lambda y.mod|do (x y) || S/NP/(S\\NP)
did:\lambda x.\lambda y.mod|do-past (x y) || S/NP,\lambda x.\lambda y.mod|do-past (x y) || S/NP/(S\\NP)
do:\lambda x.\lambda y.v|do y x || S\\NP/NP,\lambda x.\lambda y.mod|do (x y) || S/NP/(S\\NP)
does:\lambda x.\lambda y.mod|do-3s (y x) || S\\NP/(S\\NP),\lambda x.\lambda y.mod|do-3s (x y) || S/NP/(S\\NP)
dropped:\lambda x.\lambda y.v|drop-past y x || S\\NP/NP
have:\lambda x.\lambda y.v|have y x || S\\NP/NP
he:pro:sub|he || NP
his:\lambda x.det:poss|his x || NP/N,pro:poss|his || NP
hurt:\lambda x.\lambda y.v|hurt-zero y x || S\\NP/NP
in:\lambda x.\lambda y.prep|in (y x) || S\\NP\\(S\\NP)/NP,\lambda x.prep|in x || S/S
is:\lambda x.\lambda y.v|equals x y || S\\NP/NP,\lambda x.\lambda y.v|hasproperty y x || S\\NP/NP
it:pro:per|it || NP
like:\lambda x.\lambda y.v|like y x || S\\NP/NP
lost:\lambda x.\lambda y.v|lose-past y x || S\\NP/NP
may:\lambda x.\lambda y.mod|may (x y) || S\\NP/(S\\NP)
missed:\lambda x.v|miss-past x || S\\NP,\lambda x.\lambda y.v|miss-past y x || S\\NP/NP
my:\lambda x.det:poss|my x || NP/N
name:n|name || N
need:\lambda x.\lambda y.v|need y x || S\\NP/NP
no:\lambda x.qn|no x || NP/N
not:\lambda x.\lambda y.not (x y) || S\\NP/(S\\NP)\(S\\NP/(S\\NP))
on:\lambda x.prep|on x || S\\NP\\(S\\NP)/NP
one:pro:indef|one || NP
pencil:n|pencil || N
say:\lambda x.\lambda y.v|say y x || S\\NP/NP
see:\lambda x.\lambda y.v|see y x || S\\NP/NP
shall:\lambda x.\lambda y.mod|shall (x y) || S\\NP/(S\\NP)
some:\lambda x.qn|some x || NP/N
that:pro:dem|that || NP,\lambda x.pro:det|that x || NP/N
the:\lambda x.det:art|the x || NP/N
they:pro:sub|they || NP
this:pro:dem|this || NP,\lambda x.pro:det|this x || NP/N
those:pro:dem|those || NP,\lambda x.pro:det|those x || NP/N
was:\lambda x.\lambda y.v|equals x y || S\\NP/NP,\lambda x.\lambda y.v|hasproperty y x || S\\NP/NP
we:pro:sub|we || NP
what:pro:int|WHAT || Swhq/Sq/NP,pro:int|WHAT || NP
who:pro:int|WHO || Swhq/Sq/NP,pro:int|WHO || NP
you:pro:per|you || NP
your:\lambda x.det:poss|your x || NP/N
\end{verbatim}

\textbf{Hagar}
\begin{verbatim}
nakon:adv|nakon || S
ze:\lambda x.v|hasproperty pro:dem|ze x || NP
?at:pro:per|?at || NP
ken:adv|ken || S
ha:\lambda x.det|ha x || NP/N
hu?:pro:per|hu? || NP
lo?:\lambda x.\lambda y.not (x y) || S\\NP/(S\\NP)\(S\\NP/(S\\NP))
bo?i:v|ba? you || S
roca:\lambda x.\lambda y.v|raca y x || S\\NP/NP
?ani:pro:per|?ani || NP
?aba?:n:prop|?aba? || NP
qxi:v|laqax you || S
od:\lambda x.qn|od x || NP/N
taim:adj|taim || S
ro?a:\lambda x.v|ra?a x || S\\NP
tistakli:v|histakel you || S
le:\lambda x.prep|le x || S\\NP
hi?:pro:per|hi? || NP
gamarnu:v|gamar you || S
hem:pro:per|hem || NP
nafal:\lambda x.v|nafal x || S\\NP
texapsi:v|xipes you || S
kaxol:adj|kaxol || S
zo?t:pro:dem|zo?t || NP
lehitra?ot:v|hitra?a you || S
tir?i:v|ra?a you || S
betuxa:\lambda x.v|hasproperty x adj|batuax || S\\NP
qar:adj|qar || S
?ima?:n:prop|?ima? || NP
?oqer:adj|?oqer || S
halak:\lambda x.v|halak x || S\\NP
glida:n|glida || N,n|glida-BARE || NP
xam:adj|xam || S
?eyn:v|?eyn you || S\\NP
boke:\lambda x.v|baka x || S\\NP
yalda:n|yeled || N,n|yeled-BARE || NP
gvina:n|gvina || N,n|gvina-BARE || NP
tisperi:v|safar you || S
tarnegol:n|tarnegol || N,n|tarnegol-BARE || NP
yes:\lambda x.v|yes x || S\\NP
?aval:\lambda x.\lambda y.v|hasproperty x y || S\\NP/NP
?or:n|?or-BARE || NP
ricpa:n|ricpa || N,n|ricpa-BARE || NP
yeled:adj|yeled || N/N
al:\lambda x.prep|al x || S\\NP
?adom:adj|?adom || S
tagidi:v|higid you || S
tasiri:v|sar you || S
cahov:adj|cahov || S
salom:n|salom || N
,n|salom-BARE || NP



\end{verbatim}

\subsection{Model Predictions}
Tables \ref{tab:LF-syncat-preds-en} and \ref{tab:LF-syncat-preds-he} show the model predictions for Adam (English) and Hagar (Hebrew) respectively. In the interests of readability, we show only those words for which the predictions are not correct, either for the LF or for the syntactic category. For those words which are correctly predicted for both, the model predictions can be read off the ground-truth annotations, as provided in Section \ref{app:annotated-lexicon}.

\begin{table}
    \centering
    \resizebox{\textwidth}{!}{
\begin{tabular}{lllrr}
\toprule
 & pred LF & pred syncat & LF correct & syncat correct \\
\midrule
'll & $\lambda x$.$\lambda y$.mod|~will ($x$ $y$) & NP & True & False \\
can & $\lambda x$.$\lambda y$.Q (mod|can ($x$ $y$)) & NP & True & False \\
d & $\lambda x$.$\lambda y$.Q (mod|do ($x$ $y$)) & NP & True & False \\
did & $\lambda x$.$\lambda y$.Q (mod|do-past ($x$ $y$)) & S\bs NP & True & False \\
does & $\lambda x$.$\lambda y$.Q (mod|do-3s ($y$ $x$)) & NP & True & False \\
in & $\lambda x$.prep|in $x$ & NP/N & True & False \\
like & $\lambda x$.$\lambda y$.v|like $y$ $x$ & NP & True & False \\
may & $\lambda x$.$\lambda y$.mod|may ($x$ $y$) & NP & True & False \\
missed & $\lambda x$.v|miss-past $x$ & NP & True & False \\
not & $\lambda x$.$\lambda y$.not ($x$ $y$) & S\bs NP/NP & True & False \\
on & $\lambda x$.prep|on $x$ & S\bs NP & True & False \\
shall & $\lambda x$.$\lambda y$.Q (mod|shall ($x$ $y$)) & NP & True & False \\
\bottomrule
\end{tabular}
}
\caption{List of all incorrect model LF and syntactic category predictions for English.}
    \label{tab:LF-syncat-preds-en}
\end{table}

\begin{table}
    \centering
    \resizebox{\textwidth}{!}{
\begin{tabular}{lllrr}
\toprule
 & pred LF & pred syncat & LF correct & syncat correct \\
\midrule
nakon & adv|nakon & NP/N & True & False \\
ken & adv|ken & NP/N & True & False \\
lo? & $\lambda x$.$\lambda y$.not ($x$ $y$) & NP & True & False \\
bo?i & v|ba? you & NP & True & False \\
roca & $\lambda x$.$\lambda y$.Q (v|raca $y$ $x$) & Sq\bs NP & True & False \\
qxi & v|laqax you & NP & True & False \\
taim & adj|taim & NP & True & False \\
ro?a & $\lambda x$.Q (v|ra?a $x$) & Sq\bs NP & True & False \\
tistakli & v|histakel you & NP & True & False \\
gamarnu & Q (v|gamar you) & NP & True & False \\
nafal & $\lambda x$.Q (v|nafal $x$) & Sq\bs NP & True & False \\
texapsi & v|xipes you & NP & True & False \\
kaxol & adj|kaxol & NP & True & False \\
lehitra?ot & v|hitra?a you & NP & True & False \\
tir?i & v|ra?a you & NP & True & False \\
betuxa & $\lambda x$.Q (v|hasproperty $x$ adj|batuax) & Sq\bs NP & True & False \\
qar & adj|qar & NP & True & False \\
?oqer & adj|?oqer & NP & True & False \\
glida & n|glida & NP & True & True \\
xam & adj|xam & S/S & True & False \\
boke & $\lambda x$.Q (v|baka $x$) & Sq\bs NP & True & False \\
yalda & n|yeled & NP & True & True \\
gvina & Q (n|gvina) & NP & True & True \\
tisperi & v|safar you & NP & True & False \\
tarnegol & n|tarnegol & NP & True & True \\
?aval & $\lambda x$.$\lambda y$.v|hasproperty $x$ $y$ & NP & True & False \\
ricpa & n|ricpa & NP & True & True \\
yeled & adj|yeled & NP & True & False \\
?adom & adj|?adom & NP & True & False \\
tagidi & v|higid you & NP & True & False \\
tasiri & v|sar you & NP & True & False \\
cahov & adj|cahov & NP & True & False \\
salom & n|salom & NP & True & True \\
\bottomrule
\end{tabular}
}
    \caption{List of all incorrect model LF and syntactic category predictions for English. Again note that, as we evaluate here, the model can get both LF and syntactic category correct individually but still get the overall prediction wrong if the two do not match.}
    \label{tab:LF-syncat-preds-he}
\end{table}

\section{Plots of Higher Numbers of Distractors} \label{app:higher-distractors-adam}
In Section~\ref{subsec:distractors}, we following \cite{abend2017bootstrapping} in reporting the trend of word-order learning in settings with 2, 4 and 6 distractor settings for English (for us, this is the Adam corpus, for \citeauthor{abend2017bootstrapping}, this was the Eve corpus. For Hagar (Hebrew), we reported just 2 distractors, because already this was too much for the model to learn word order effectively, for the reasons discussed in Section~\ref{subsec:distractors}. Here, we report high number of distractors for Adam: 8, 10 and 12, which shows further robustness to noise in the LFs when learning English word order. For 8 and 10 distractors, SVO is still learnt confidently and relatively smoothly. For 12 distractors, it takes much longer before SVO starts to dominate, but by the end, the learner has still quite firmly acquired SVO. Note that the differences in relative probability between the six orders are more significant towards the end of training, because by that time that model has seen more data and so it would take more data again for it to change its mind. Formally, the denominators in the Dirichlet processes have become large. Therefore, the spike of SVO at the end is more significant than the spike of OSV at the beginning.

\begin{figure}
    \centering
    \includegraphics[width=0.65\textwidth]{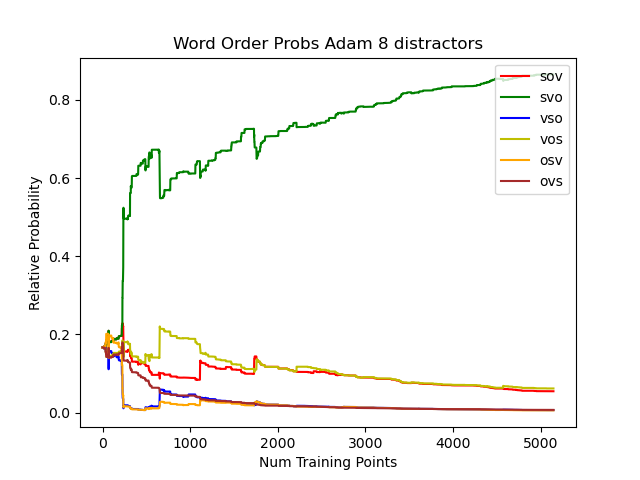}
    \includegraphics[width=0.65\textwidth]{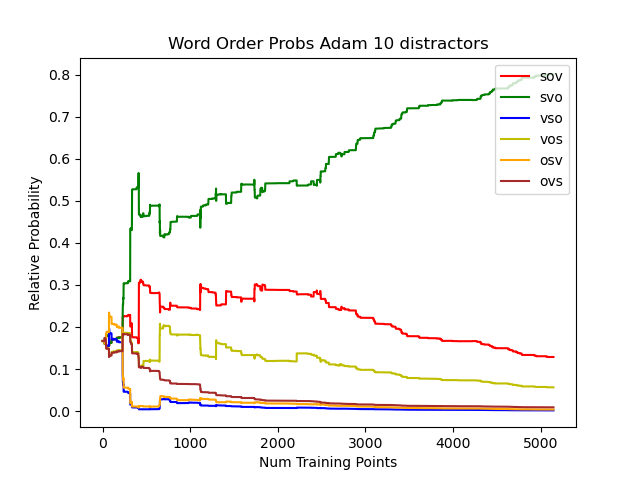}
    \includegraphics[width=0.65\textwidth]{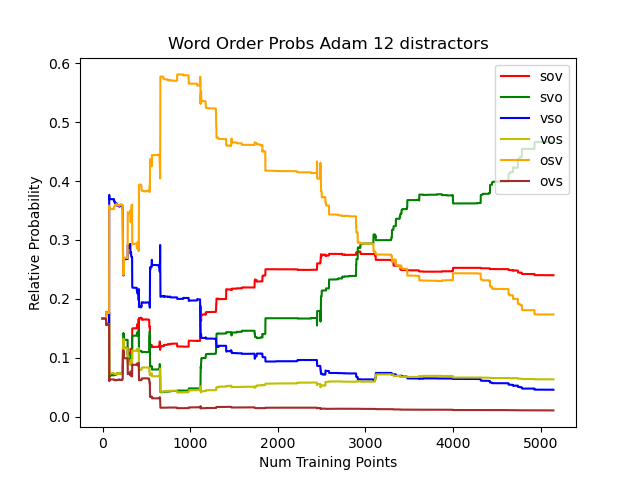}
    \caption{Word-order learning in Adam (English) with higher numbers of distractor logical forms.}
    \label{fig:adam-higher-distractors}
\end{figure}

\section{Plots vs Number of Tokens}
\begin{figure}
    \centering
    \includegraphics[width=0.8\textwidth]{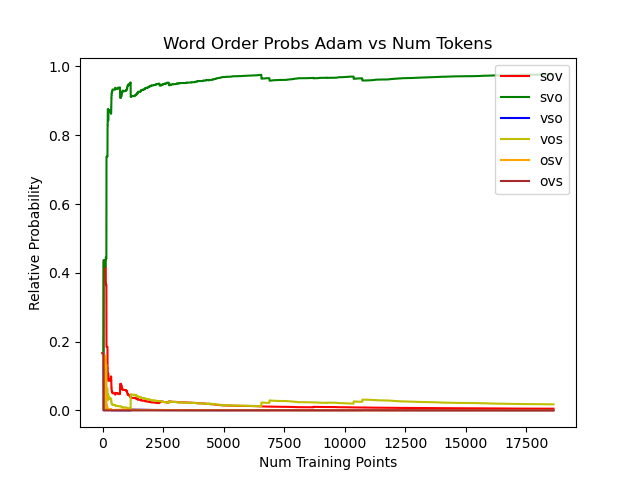}
    \includegraphics[width=0.8\textwidth]{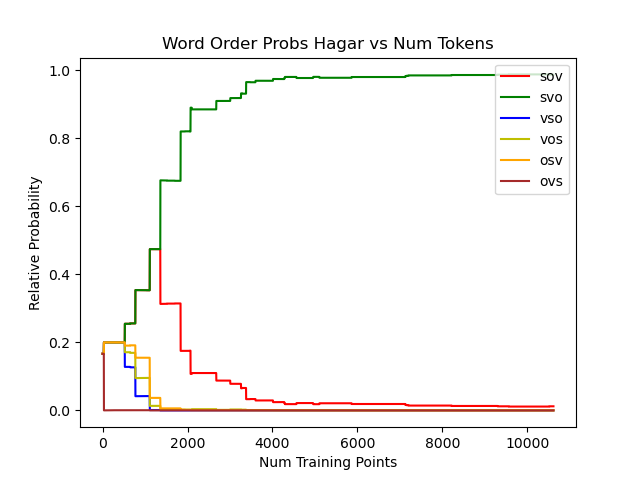}
    \caption{Plot of the relative probability of the six word orders as a function of the number of tokens the model as seen during training.}
    \label{fig:word-order-probs-vs-n-toks}
\end{figure}

Because the average number of tokens differs between the two corpora, one may also want to consider how learning develops as a function of the number of tokens seen rather than the number of utterances. This is shown in Figure \ref{fig:word-order-probs-vs-n-toks}.

\end{document}